\documentclass[runningheads]{llncs}
\usepackage{amsmath} % I added this package
\usepackage[T1]{fontenc}
\usepackage{graphicx}
\usepackage{xcolor}
\usepackage{multirow}
\usepackage{booktabs}
\usepackage{cite}
\usepackage[acronym]{glossaries}
\makeglossaries

\newacronym{imo}{IMO}{Independently Moving Object}
\newacronym{foe}{FOE}{Focus of Expansion}
\newacronym{iwe}{IWE}{Image of Warped Events}

\begin{document}
\title{GeoIMO: Geometry-Driven Independent Motion Classification for Event Cameras}
\titlerunning{GeoIMO: Geometry-Driven IMO Classification}
\author{Anil Bayram Gogebakan\inst{1} \and
Filippo Marostica\inst{1} \and
Alessio Caviglia\inst{1} \and
Alessandro Savino\inst{1} \and
Stefano Di Carlo\inst{1}}

\authorrunning{A. B. Gogebakan et al.}
\institute{Politecnico di Torino, Turin, Italy\\
\email{\{anil.gogebakan, filippo.marostica, alessio.caviglia, alessandro.savino, stefano.dicarlo\}@polito.it}}
\maketitle              % typeset the header of the contribution
\begin{abstract}
Existing automotive event datasets rely on appearance-based annotations from frame pipelines, making them poorly suited for motion-aware event perception. We present a geometry-driven, annotation-free framework that classifies detected objects as static or independently moving by exploiting ego-motion structure directly from the event stream. A Focus of Expansion model with yaw compensation estimates global background motion, while objects are labeled as moving when local motion deviates from this prediction, as quantified by a scale-invariant residual. Temporal stabilization improves robustness across consecutive event windows. The method requires no learning, no manual motion labels, and works with any input bounding boxes. Experiments on MVSEC and the Prophesee 1 Megapixel Automotive Detection dataset demonstrate consistent performance across diverse driving scenarios, with yaw compensation improving results during turns and a simple translational local model offering a favorable accuracy-efficiency trade-off.
\footnote{The code is publicly available at \url{https://github.com/smilies-polito/GeoIMO}.}

\end{abstract}

\section{Introduction}
\label{sec:intro}

Autonomous driving systems in complex, dynamic environments must reliably separate stationary scene elements, ego-motion, and \glspl{imo}. This is essential for safety-critical tasks such as collision avoidance, trajectory prediction, and motion planning~\cite{yurtseverSurveyAutonomousDriving2020}. Traditional vision-based systems are limited by appearance-based object detection, which treats parked vehicles and moving traffic alike. In safety-critical settings, this motion-agnostic approach can cause overly conservative behavior or fail to prioritize dynamic threats. Overcoming this limitation requires perception systems that capture motion at high temporal resolution.

Event cameras meet this need by replacing frame-based capture with asynchronous, pixel-level event generation~\cite{gallegoEventBasedVisionSurvey2022,lichtsteiner128times1281202008}, offering microsecond temporal resolution, inherent motion sensitivity, low latency, and robustness to motion blur and high dynamic range scenes. However, their adoption remains limited by the need for specialized algorithms and data representations, immature processing pipelines, and the scarcity of large-scale annotated datasets for event-based perception~\cite{gallegoEventBasedVisionSurvey2022,chenEventBasedNeuromorphicVision2020}.

A key obstacle is that existing automotive datasets lack bounding boxes reflecting object motion state: most annotations derive from RGB pipelines and treat static and moving objects alike~\cite{perotLearningDetectObjects2020,zhuMultivehicleStereoEvent2018}. This appearance-based labeling creates a modality mismatch that limits the ability to distinguish \glspl{imo} from static scene elements, and makes such datasets unsuitable as ground truth for methods that exploit the motion sensitivity of event cameras.

Existing approaches to motion understanding in event streams suffer from significant limitations. Supervised methods require costly per-object manual annotation of motion state. Dense optical flow estimation is computationally expensive, sensitive to noise in sparse event data, and requires additional ego-motion compensation before object motion can be isolated \cite{stoffregenEventBasedMotionSegmentation2019}. Tracking-based methods depend on temporal association and are fragile in the presence of occlusions, clutter, and abrupt motion changes. Critically, none of these approaches exploit the geometric structure of ego-motion already latent in the event stream in a lightweight, annotation-free manner.

This work proposes a geometry-driven motion-based filtering framework that leverages parametric ego-motion models to classify detected objects as static or independently moving. The core insight is that vehicle ego-motion induces a globally consistent, predictable motion pattern across the event stream. By estimating this global pattern and comparing it with the local motion observed within detected object regions, objects that deviate from expected ego-motion can be identified as independently moving. The framework uses the \gls{foe}~\cite{longuet-higginsInterpretationMovingRetinal1980}, a compact geometric representation of translational ego-motion derived from optical flow, to characterize background motion without requiring dense flow computation. Motion consistency of events within each bounding box is then assessed relative to the estimated \gls{foe}, enabling classification of \glspl{imo} without any learned components or manual motion annotation. Importantly, the framework operates on any set of bounding boxes, whether provided by an existing dataset or generated on the fly by applying an off-the-shelf frame-based detector to co-recorded imagery, and produces motion-consistent labels suitable for training and evaluating event-based perception systems. 

This approach bridges the gap between frame-based annotations and event-based perception, enabling the construction of motion-consistent object representations better aligned with the sensing characteristics of event cameras. Beyond dataset annotation, the proposed framework provides a foundation for downstream tasks, including motion segmentation, moving-object detection, neuromorphic perception, and temporally-aware tracking.

\section{Background}

\subsection{Event Cameras, Event-Based Automotive Datasets, and the Annotation Gap}

Event cameras are bio-inspired sensors that replace frame-based capture with asynchronous event generation~\cite{lichtsteiner128times1281202008,gallegoEventBasedVisionSurvey2022}. Each pixel emits an event $(x, y, t, p)$ when its log-brightness changes beyond a threshold, where $(x,y)$ is the pixel location, $t$ the microsecond timestamp, and $p \in \{+1,-1\}$ the polarity. This produces sparse, high-temporal-resolution data that naturally encodes motion: static scene elements generate no events, while moving edges produce dense activity. These properties make event cameras well-suited for motion-aware perception in dynamic driving environments~\cite{chenEventBasedNeuromorphicVision2020,gallegoEventBasedVisionSurvey2022}.

Public datasets have been central to progress in event-based autonomous driving, with the two main object-level benchmarks being Prophesee 1Megapixel Automotive Detection~\cite{perotLearningDetectObjects2020} and MVSEC~\cite{zhuMultivehicleStereoEvent2018}. Prophesee provides RGB-derived bounding boxes transferred to the event domain via homography, while MVSEC offers synchronized event and grayscale data but no native bounding-box annotations. In both datasets, labels reflect appearance rather than motion state, so static objects may still be marked as foreground, creating a modality mismatch for methods that rely on the motion sensitivity of event cameras.

\subsection{Motion Compensation and Contrast Maximization}

A principled way to interpret event streams is \emph{motion compensation}: events are warped using candidate motion parameters, and their alignment is evaluated to estimate the underlying motion. Here, the motion parameters describe the assumed image motion, for example, a horizontal and vertical velocity $(v_x, v_y)$, or a more general parametric motion model such as optical flow or ego-motion parameters. For a given motion field, each event at location $(x_i, y_i)$ and time $t_i$ is shifted to a reference time $t_\text{ref}$, and accumulated into an \gls{iwe}. If the motion parameters correctly describe the scene dynamics, events from the same physical edge align spatially and form sharp structures in the \gls{iwe}; otherwise, the \gls{iwe} becomes blurred.

This idea is formalized through \emph{contrast maximization}~\cite{stoffregenEventCamerasContrast2019}, which measures alignment quality by the spatial variance of the \gls{iwe}. Motion is then estimated by maximizing this contrast over the candidate parameter space, without explicit supervision. Beyond ego-motion estimation, \emph{contrast maximization} has also been used for motion segmentation, separating the event stream into independently moving components \cite{stoffregenEventBasedMotionSegmentation2019}. However, Stoffregen et al.~\cite{stoffregenEventBasedMotionSegmentation2019} lack an explicit geometric ego-motion model, making it vulnerable to yaw-induced flow corruption during turns, susceptible to background cluster contamination in dense automotive scenes, and dependent on a fixed pre-specified number of objects, limitations that directly motivate the geometry-driven approach proposed here. The proposed framework builds on the same contrast maximization objective but applies it separately to ego-motion and object-level estimation under an explicit geometric parameterization, avoiding joint iterative clustering entirely.

\subsection{Optical Flow, Ego-Motion, and the Focus of Expansion}

Optical flow describes the apparent 2D pixel velocity induced by relative motion between the camera and the scene~\cite{hornDeterminingOpticalFlow1981}. For event cameras, it can be estimated directly from event streams~\cite{zhuEVFlowNetSelfSupervisedOptical2018}, linking raw event data to higher-level motion understanding.
Under pure camera translation in a static scene, the optical flow field is radially symmetric: all vectors point away from a single image location, the \gls{foe}~\cite{longuet-higginsInterpretationMovingRetinal1980}. The \gls{foe} is the projection of the translation direction onto the image plane, and the flow magnitude increases with distance from it. Since driving is often dominated by forward motion, the \gls{foe} offers a compact and efficient representation of background ego-motion.

Yaw rotation, caused by steering, adds an approximately uniform horizontal flow component that radial expansion alone cannot explain. As a result, the apparent \gls{foe} is shifted, and a pure translation model becomes biased during turns. Prior work therefore extends the \gls{foe} model with an explicit rotational term~\cite{heegerSubspaceMethodsRecovering1992, nagataFOEbasedRegularizationOptical2019}, enabling more accurate ego-motion estimation across a wider range of driving maneuvers. This combined parameterization forms the geometric basis of our method.

\subsection{Object-Level Motion Reasoning in Event Streams}

Despite progress in optical flow and ego-motion estimation, connecting pixel- or event-level motion cues to object-level representations remains an open problem. The DOTIE framework shows that spiking neural networks can isolate independently moving objects directly from event streams~\cite{nagarajDOTIEDetectingObjects2023}, highlighting the potential of event-driven detection. However, its evaluation still relies on bounding boxes derived from frame-based annotations, inheriting the same modality mismatch.
Crucially, none of these methods exploit the geometric structure of ego-motion already present in the event stream in a lightweight, annotation-free way. This gap motivates methods that explicitly connect ego-motion estimation with object-level motion classification in event-based perception.

\section{Methodology}

The proposed framework classifies detected objects as \textit{static} or \textit{moving} using event-based motion analysis. It assumes bounding boxes are already available and focuses only on estimating the motion state of each object. The input is an event stream $E = \{x, y, t, p\}$ and a set of bounding boxes $B_t$ for each temporal window $\Delta t$, and the output is a binary motion label for each box.

\begin{figure}[h]
    \centering
    \includegraphics[width=0.7\textwidth]{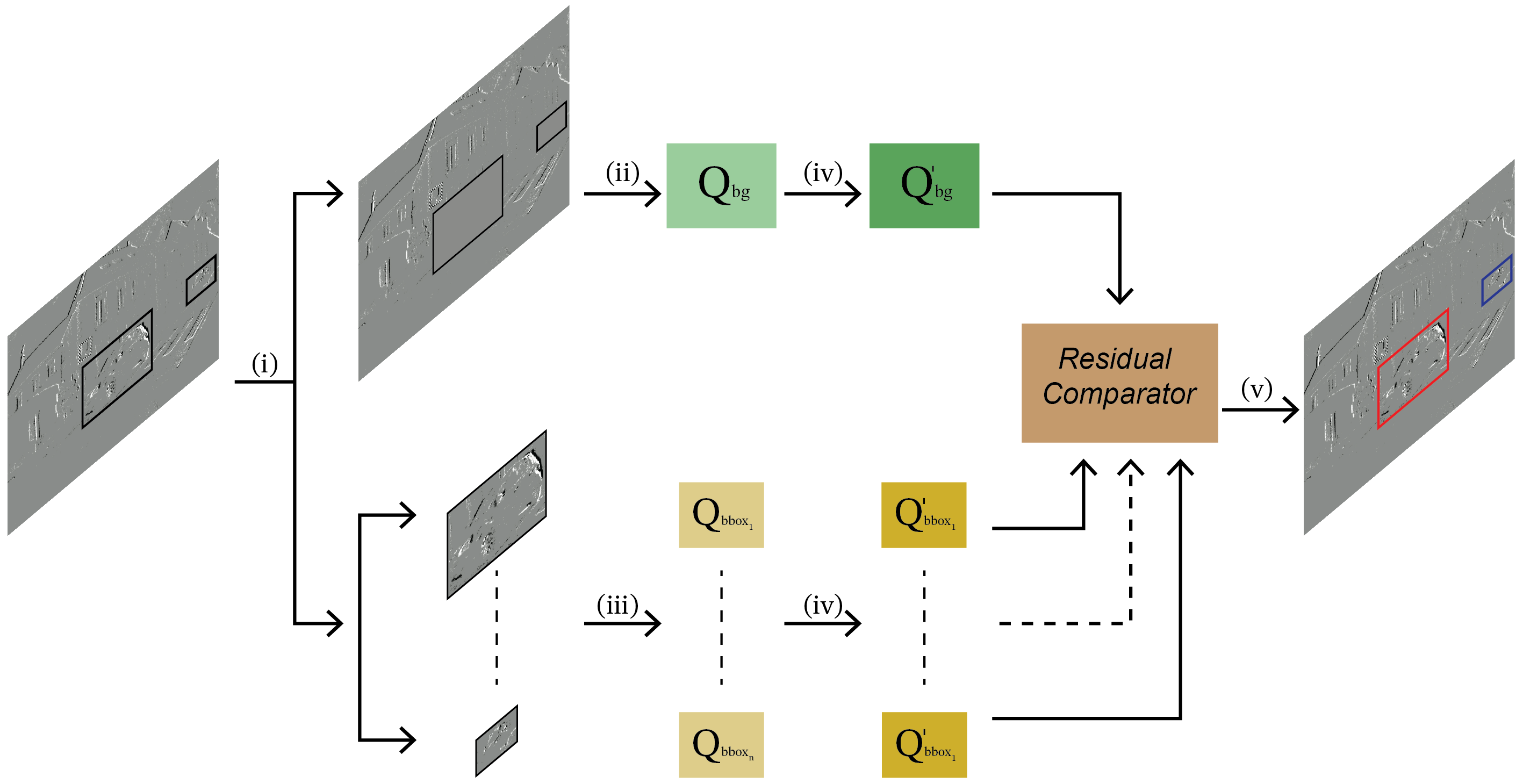}
    \caption{Overview of the proposed pipeline.}
    \label{fig:overview_figure}
\end{figure}

The pipeline consists of five stages (see Figure~\ref{fig:overview_figure}): (i) background event selection, (ii) global ego-motion estimation via contrast maximization, (iii) local motion estimation within each bounding box, (iv) temporal stabilization of global and local estimates, and (v) residual-based static/moving classification. 

\subsection{Global Ego-Motion Estimation}

\paragraph{Background event selection.}
Reliable ego-motion estimation requires events originating exclusively from static scene structures. Events located inside bounding boxes classified as non-static in previous windows are therefore excluded, as their motion patterns do not correspond to the global camera-induced motion and would otherwise bias the estimation. The remaining events are treated as background and used to estimate global ego-motion.
In the first temporal window, no prior classification is available; all bounding box events are therefore included in the background set.

\paragraph{Ego-motion model.}
Rather than estimating dense optical flow, we adopt a low-dimensional parametric model that captures the dominant effects of camera translation and rotation. Under pure forward translation, the optical flow field is radial and parameterized by the \gls{foe} $(x_{foe}, y_{foe})$~\cite{longuet-higginsInterpretationMovingRetinal1980} and an expansion coefficient $s$, yielding image-plane velocities $v_x = s(x - x_{foe})$ and $v_y = s(y - y_{foe})$.

In practice, vehicle turning introduces a yaw component that displaces the apparent \gls{foe} and cannot be captured by radial expansion alone. The model is therefore extended with a yaw parameter $\omega_y$~\cite{heegerSubspaceMethodsRecovering1992, nagataFOEbasedRegularizationOptical2019}, giving the full parameter set $\theta = (x_{foe}, y_{foe}, s, \omega_y)$. Defining centered coordinates $\bar{x} = x - c_x$ and $\bar{y} = y - c_y$ relative to the principal point $(c_x, c_y)$, the total velocity field becomes

\[
v_x = s(x - x_{foe}) - \left(f + \frac{\bar{x}^2}{f}\right)\omega_y,
\qquad
v_y = s(y - y_{foe}) - \frac{\bar{x}\bar{y}}{f}\,\omega_y,
\]

where $f$ is the focal length.

Given a candidate parameter set $\theta$, each event at location $(x_i, y_i)$ and time $t_i$ is warped toward a reference timestamp $t_\text{ref}$ according to the estimated velocity field:
\[
(x_i', y_i') = (x_i, y_i) + \bigl(v_x(x_i,y_i),\; v_y(x_i,y_i)\bigr)(t_\text{ref} - t_i).
\]
The warped events are accumulated into an \gls{iwe}, whose sharpness reflects the quality of the motion estimate. Alignment quality is quantified by the spatial variance of the \gls{iwe}, following the contrast maximization framework of~\cite{stoffregenEventCamerasContrast2019}. To prevent cancellation between opposite polarities, the contrast is computed separately for positive and negative events:
\[
C(\theta) = C^{+}(\theta) + C^{-}(\theta),
\qquad
C^{\pm}(\theta) = \frac{1}{N_p}\sum_{x,y}\bigl(I^{\pm}(x,y;\theta) - \bar{I}^{\pm}(\theta)\bigr)^2,
\]
where $N_p$ is the number of pixels and $\bar{I}^{\pm}$ the mean of the corresponding \gls{iwe}. The optimal ego-motion parameters are obtained as $\theta^* = \arg\max_\theta C(\theta)$.

\subsection{Local Bounding-Box Motion Estimation}

For each temporal window, events within each bounding box are extracted and
used to estimate local object motion independently of the global ego-motion
estimate. To avoid unreliable estimates from small or weakly activated regions,
a box $b$ is processed only if $n_b \geq \lfloor 100\rho \rfloor$ and
$A_b \geq \lfloor 150\rho \rfloor$, where $n_b$ is the number of events in the
box, $A_b$ its area, and $\rho=W_sH_s/(346\cdot260)$ scales the thresholds
with the sensor resolution. Boxes failing either criterion are assigned the
static label by default, since there is insufficient event evidence to reliably
infer independent motion.

Two motion models are considered: a radial model
$(x^b_{\mathrm{foe}}, y^b_{\mathrm{foe}}, s^b)$, which captures expansion or
contraction patterns, and a uniform translational model $(v_x,v_y)$, which
approximates lateral image-plane motion. In both cases, parameters are estimated
by maximizing contrast over events inside the bounding box.

The estimated parameters are converted into a single observed velocity
$\mathbf{v}_{\mathrm{obs}}=(v_x(x_b,y_b),v_y(x_b,y_b))$ at the box center
$(x_b,y_b)$. For the translational model, this is directly the estimated
$(v_x,v_y)$; for the radial model, it is obtained by evaluating the radial field
at the box center.

\subsection{Temporal Stabilization}

Motion estimates from individual temporal windows can be unstable due to variations in event density and imperfect contrast-maximization convergence. To reduce this variability, both global and local estimates are temporally stabilized across consecutive windows.

For ego-motion, the four parameters $(x_{foe}, y_{foe}, s, \omega_y)$ are temporally filtered using a lightweight Kalman-based formulation, applied independently to each component with different dynamics. The expansion coefficient $s$ is allowed to change more quickly to follow variations in forward speed, while the FOE coordinates, especially the vertical one, are more strongly stabilized because they vary more slowly in typical driving.

For each bounding box, the image-plane velocity $\mathbf{v}_{obs}$ is stabilized in both Cartesian and polar form. Separating magnitude and direction improves robustness, since direction becomes unreliable when velocity is small and should not affect the magnitude estimate. Each window is weighted by its contrast score: high-contrast estimates contribute more, while low-contrast, poorly aligned measurements are down-weighted. The \gls{foe} search bounds expand proportionally to the estimated yaw rate, maintaining robustness during turns.

\subsection{Motion Consistency Criterion and Classification}

Given the temporally stabilized estimates, each bounding box is classified by comparing the motion predicted by ego-motion with the locally observed motion. For a bounding box centered at $(x_b, y_b)$, the estimated ego-motion field yields an expected velocity $\mathbf{v}_{ego}(x_b, y_b)$ for a nominally static point at that location. The discrepancy with respect to the observed velocity $\mathbf{v}_{obs}$ is quantified by the scale-invariant residual

\[
r =
\frac{
\left\| \mathbf{v}_{obs} - \mathbf{v}_{ego} \right\|
}{
\max\!\left(\|\mathbf{v}_{obs}\|,\|\mathbf{v}_{ego}\|,\epsilon\right)
},
\]

where $\epsilon$ prevents numerical instability when both velocities are near zero. Normalization by the larger of the two magnitudes ensures that $r$ is invariant to the overall motion scale, enabling a single threshold $\tau$ to operate consistently across different driving speeds and object distances. Each object is labeled static if $r \leq \tau$ and moving if $r > \tau$.

\section{Experiments and Results}

\subsection{Experimental Setup}

Experiments are conducted on two widely used event-based automotive datasets: MVSEC~\cite{zhuMultivehicleStereoEvent2018} and the Prophesee dataset~\cite{perotLearningDetectObjects2020}. MVSEC provides synchronized event streams and grayscale images but does not include native bounding-box annotations. Therefore, object bounding boxes are obtained by applying a YOLOv3 detector on the grayscale frames. In contrast, the Prophesee dataset provides bounding box annotations directly. In this section, we first present results on MVSEC, followed by a more detailed analysis on a selected subset of the Prophesee dataset.

Since existing annotations do not distinguish between static and moving objects, all bounding boxes are manually annotated as either \textit{static} or \textit{moving}. These labels serve as ground truth for evaluating motion classification performance. Manual annotation ensures motion-consistent labels aligned with the event data.
The proposed method performs binary classification at the bounding-box level, assigning each object a \textit{static} or \textit{moving} label. Predicted and ground-truth boxes are matched based on spatial and temporal proximity to ensure consistent comparison over time. Evaluation is then performed only on matched pairs, isolating motion-classification performance from detection errors.
Since each error simultaneously counts as a false positive for one class and a false negative for the other, the total numbers of false positives and false negatives are equal, making micro-precision, micro-recall, and micro-F1 identical. For class-balanced evaluation, we therefore use per-class F1 and macro F1.

\subsection{Motion Classification Results on MVSEC}

We evaluate the proposed framework on the MVSEC dataset using the \textit{outdoor\_day2} sequence. Bounding boxes are generated by applying a YOLOv3 detector on the grayscale frames and subsequently annotated with motion labels. The resulting dataset is highly imbalanced, with approximately 88.5\% of the bounding boxes corresponding to static objects and only 11.5\% to moving ones. This imbalance is an important factor when interpreting classification performance, particularly for the moving class.

\begin{table}[t]
\centering
\caption{Motion classification performance on the MVSEC \textit{outdoor\_day2} sequence.}
\label{tab:mvsec_results}
\begin{tabular}{lccccc}
\hline
Yaw & BBox Model & F1-score & Moving F1 & Static F1 & Macro F1 \\
\hline
False & Translational & 0.6865 & 0.2700 & 0.8004 & 0.5352 \\
True  & Translational & 0.7192 & 0.2911 & 0.8250 & 0.5580 \\
False & Radial        & 0.6955 & 0.2788 & 0.8070 & 0.5429 \\
True  & Radial        & 0.7211 & 0.2910 & 0.8264 & 0.5587 \\
\hline
\end{tabular}
\end{table}

Table~\ref{tab:mvsec_results} reports motion classification performance under different motion modeling configurations. We compare the impact of incorporating yaw rotation into the ego-motion model and of different local motion models within bounding boxes.
The results show a consistent gap between the moving and static classes. While the static class achieves high F1-scores (above 0.80 in all configurations), the moving class remains significantly lower (around 0.27–0.29). This is partly due to the strong class imbalance, which biases the model toward static predictions. In addition, the \textit{outdoor\_day2} sequence contains multiple hard turns, where accurate FOE estimation becomes challenging. Errors in global motion estimation during and shortly after these turns lead to incorrect motion compensation, which particularly affects the moving class.
In terms of the motion model, enabling yaw consistently improves performance across both bounding box models. This is expected, as rotational motion becomes significant in turning scenarios. Including the yaw parameter allows the model to better capture these dynamics, resulting in improved motion compensation and classification accuracy.
Comparing the bounding box motion models, the radial formulation yields slightly higher performance than the translational one across all configurations. However, the improvement is marginal, indicating that the additional modeling complexity does not translate into a substantial gain in classification accuracy. While the radial model is theoretically more expressive, the limited spatial extent of bounding boxes and the sparsity of event data restrict its practical advantage.
Overall, the best performance is achieved with the radial model and yaw enabled. Nevertheless, the small performance gap between radial and translational models suggests that simpler motion formulations remain competitive, particularly in terms of computational efficiency.

\subsection{Motion Classification on Prophesee Subset}

We further evaluate the proposed framework on a selected subset of the Prophesee dataset. The subset includes multiple driving scenarios with varying motion characteristics, enabling a more detailed analysis of the impact of motion modeling choices.

\subsubsection{Effect of Yaw in Ego-Motion Modeling}

\begin{figure}[tp]
    \centering
    \includegraphics[width=\linewidth]{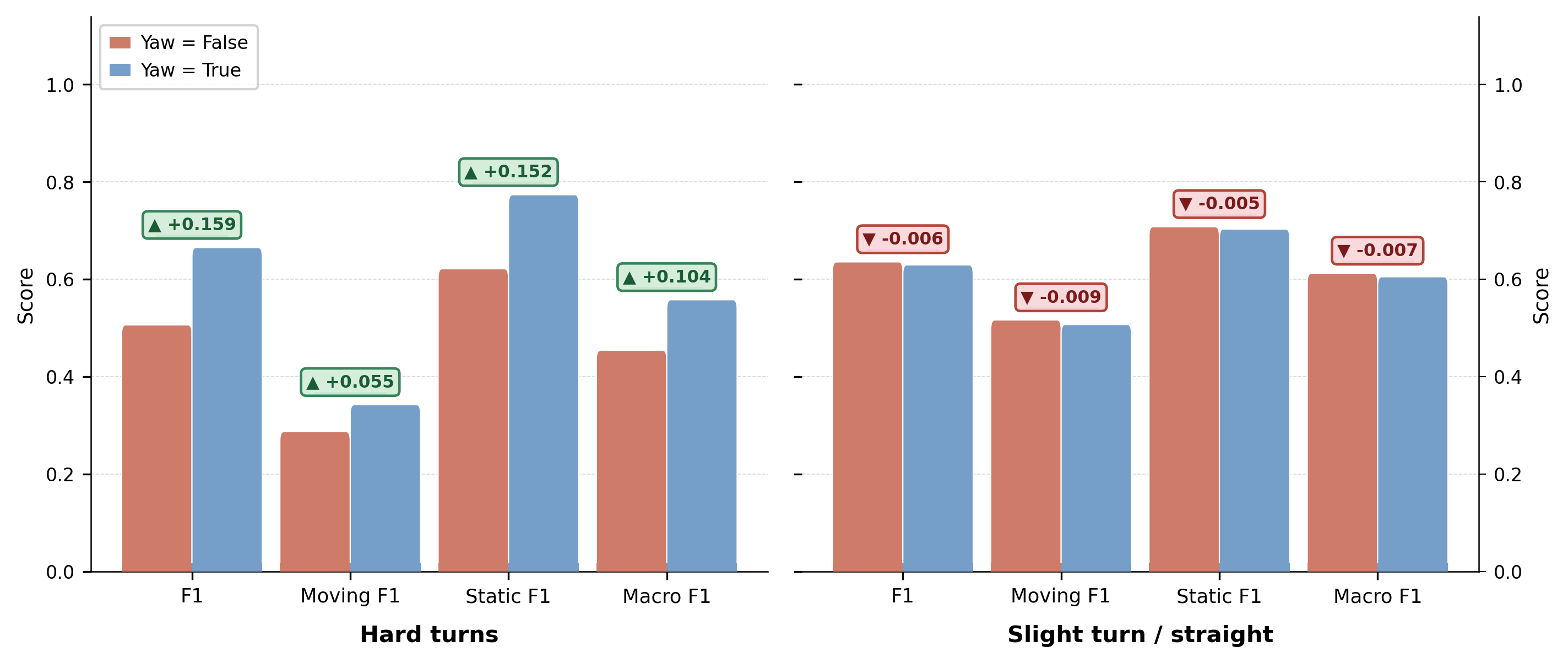}
    \caption{Classification performance with and without yaw compensation across two driving scenarios from Prophesee dataset. Badges report the signed difference $\Delta = \text{yaw}_\text{true} - \text{yaw}_\text{false}$. Yaw compensation yields substantial gains in hard-turn sequences, while its effect is negligible under near-straight ego-motion.}
    \label{fig:yaw_effect_combined}
\end{figure}

We first analyze the impact of incorporating yaw rotation in the global motion model. Figure~\ref{fig:yaw_effect_combined} reports classification performance with and without yaw across two driving scenarios, with all other parameters held fixed. The annotated badges show the signed difference $\Delta = \text{yaw}_\text{true} - \text{yaw}_\text{false}$ for each metric, highlighting the contribution of yaw compensation directly on the figure.

The results show that incorporating yaw significantly improves performance in scenarios involving strong camera rotations, such as hard turns. In contrast, for near-straight motion, including yaw does not provide measurable benefits and can slightly degrade performance. This indicates that the additional model complexity introduced by yaw is only beneficial when the underlying motion justifies it.

\subsubsection{Effect of Bounding Box Motion Model}

We next analyze the impact of the local motion model used within bounding boxes. Table~\ref{tab:bbox_motion_video_deltas} compares the radial and translational models in terms of both classification performance and runtime.

\begin{table}[t]
\centering
\caption{Per-sequence difference between radial and translational bounding box motion models (radial $-$ translational).}
\label{tab:bbox_motion_video_deltas}
\begin{tabular}{lcc}
\hline
\textbf{Sequence} & $\Delta$F1 & $\Delta$Time (s/prediction)\\
\hline
Scene 1 & 0.025 & 0.30 \\
Scene 2 & -0.034 & 0.79 \\
Scene 3 & -0.014 & 0.79 \\
Scene 4 & -0.016 & 0.35 \\
\hline
\end{tabular}
\end{table}

The radial model, while more expressive, does not consistently outperform the simpler translational model in terms of F1-score. In several sequences, the translational model achieves comparable or even better performance. At the same time, the computational cost of the radial model is significantly higher.

A likely explanation for this behavior is the limited spatial extent of bounding
boxes. Since each bounding box contains a relatively small number of events, the
additional degrees of freedom introduced by the radial model cannot be reliably
estimated. Furthermore, local event distributions can be affected by sensor noise,
event-rate variability, and sparse activity within small bounding boxes, which
can further degrade the stability of more complex motion models. As a result,
the radial model provides no reliable benefit in practice despite its higher
theoretical expressiveness.

These findings suggest that the translational model provides a better trade-off between accuracy and efficiency for object-level motion estimation.

\subsubsection{Global vs.\ Scenario-Specific Parameter Selection}

Finally, we compare two parameter selection strategies: a single global configuration applied across all sequences, and a scenario-specific configuration optimized for each sequence. The results are summarized in Table~\ref{tab:strategy_comparison}.

\begin{table}[t]
\centering
\caption{Comparison between a global parameter configuration and per-sequence optimized configurations.}
\label{tab:strategy_comparison}
\begin{tabular}{lcccc}
\hline
\textbf{Strategy} & \textbf{F1-score} & \textbf{Moving F1} & \textbf{Static F1} & \textbf{Macro F1} \\
\hline
Global Best (Single Set) & 0.651 & 0.393 & 0.749 & 0.571 \\
Per-Video Best           & 0.701 & 0.435 & 0.792 & 0.614 \\
\hline
\end{tabular}
\end{table}

Using a single global configuration already yields strong performance across different scenarios, demonstrating the robustness of the proposed framework. However, further improvements can be achieved by adapting the model parameters to specific sequences. This indicates that different motion conditions benefit from different modeling assumptions, and suggests that scenario-aware or adaptive parameter selection could further enhance performance.

\subsubsection{Qualitative Analysis}

\begin{figure}[t]
\centering
\includegraphics[width=0.48\linewidth]{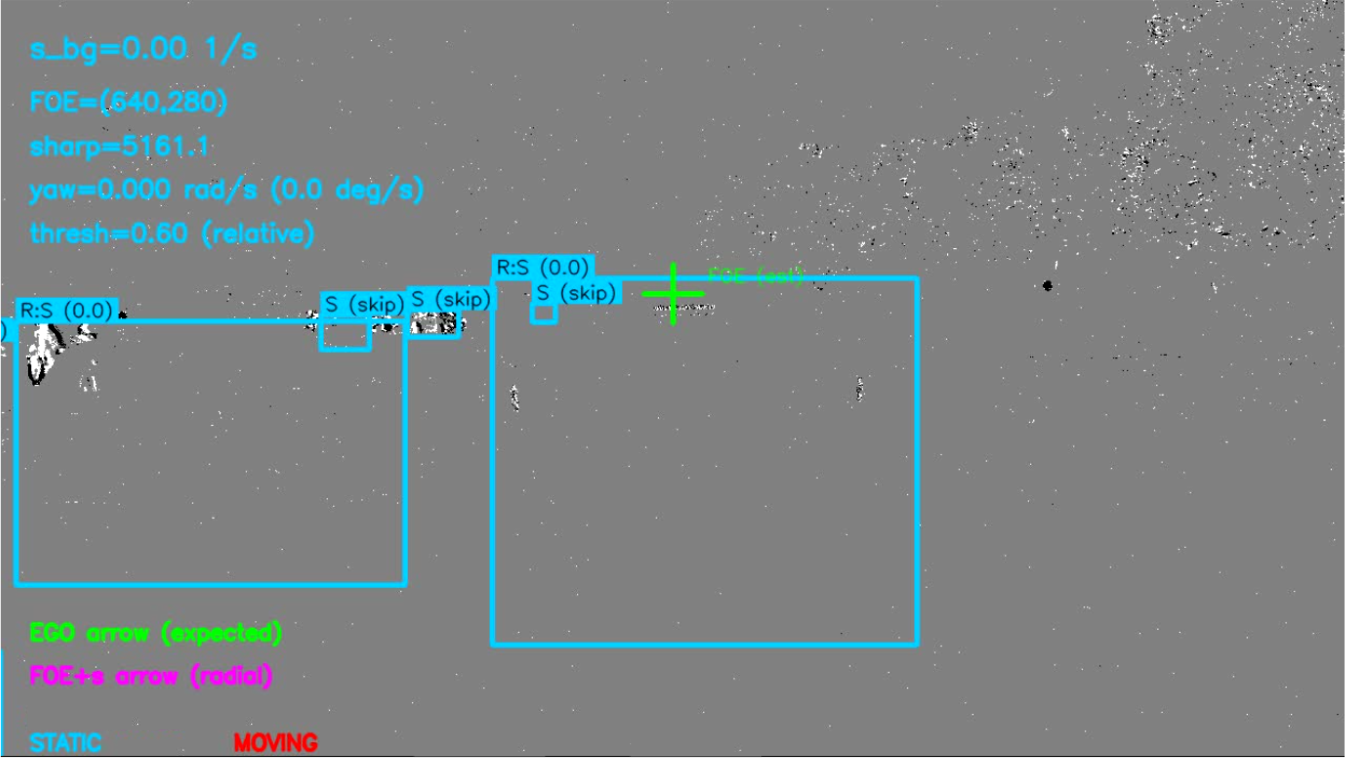}
\hfill
\includegraphics[width=0.48\linewidth]{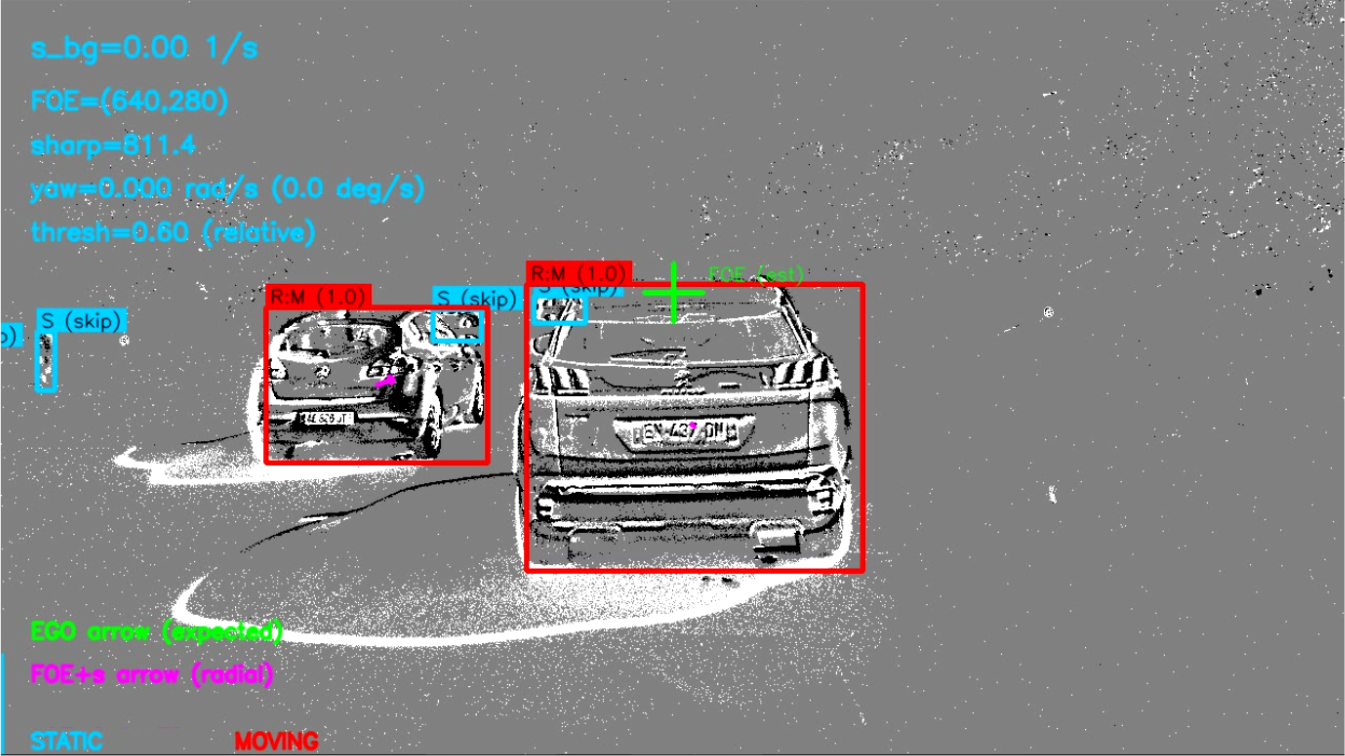}
\caption{Event activity in bounding boxes before and after motion onset.}
\label{fig:static_vs_moving_events}
\end{figure}

\begin{figure}[t]
    \centering
    \includegraphics[width=0.48\linewidth]{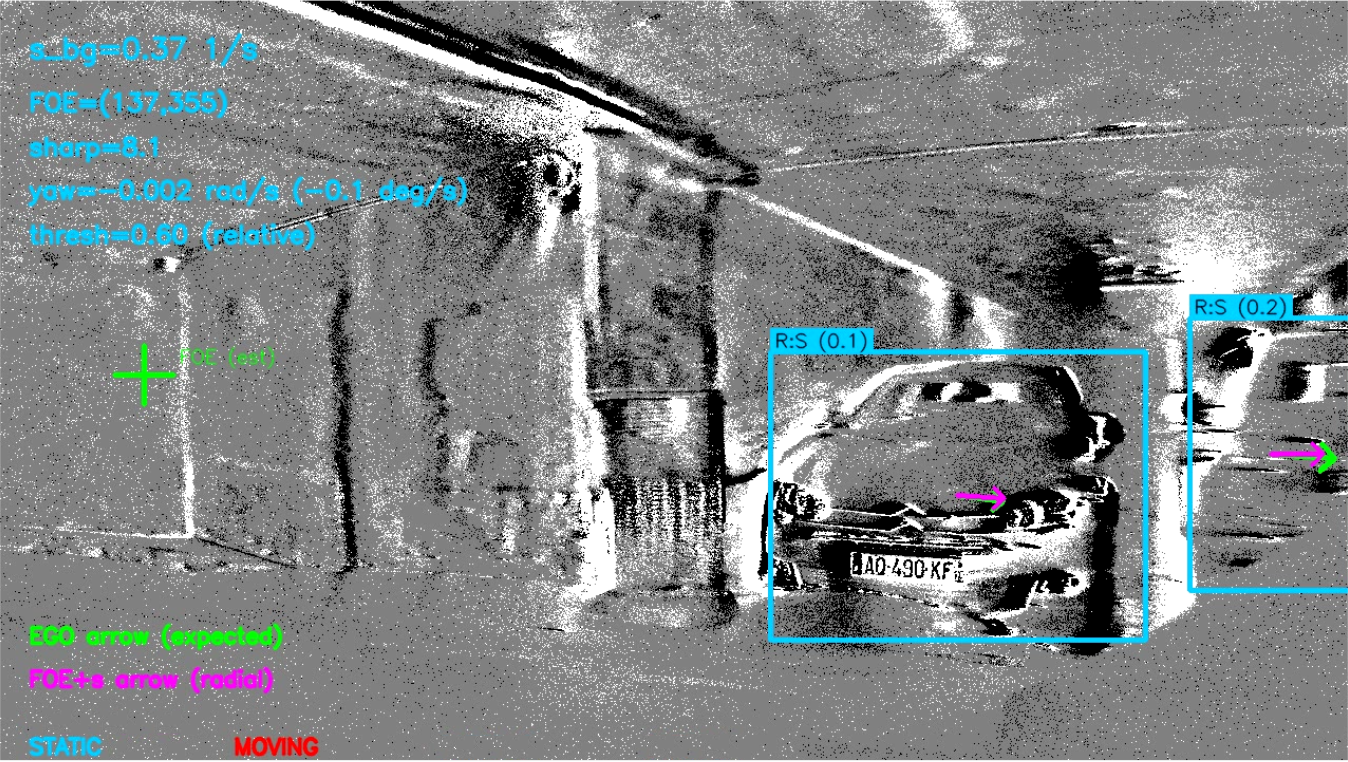}
    \caption{Example of ego-motion during a hard turn, showing the corresponding shift of the \gls{foe}.}
    \label{fig:single_ex}
\end{figure}

We provide qualitative examples that highlight the mismatch between frame-based detections and event-based motion cues, which motivate our approach.
Figure~\ref{fig:static_vs_moving_events} shows a traffic scene before and after motion onset. While bounding boxes are present even when the scene is static, almost no events are generated inside them. Once motion begins, event activity rises sharply within the same regions. This shows that frame-based detectors capture object presence, whereas event data reflects scene dynamics, motivating motion-aware filtering of bounding boxes.
Figure~\ref{fig:single_ex} presents a hard-turn example. As the ego vehicle rotates left, the estimated \gls{foe} shifts toward the left side of the image, consistent with the geometric motion model. This helps explain the performance gains obtained by incorporating yaw.

\section{Conclusion}

This work presented a geometry-driven framework for classifying bounding boxes as static or moving from event-based motion cues. By combining parametric ego-motion estimation with contrast maximization, the method operates without supervision or explicit motion labels, addressing the mismatch between frame-based annotations and event-driven sensing.
Experiments on MVSEC and Prophesee show consistent performance across diverse driving scenarios. Yaw modeling improves results during turns, while simple translational models offer a promising trade-off between accuracy and efficiency. A single global parameter setting already performs competitively, with scenario-specific tuning providing further gains.
Overall, the approach bridges appearance-based annotations and motion-consistent event perception, offering a lightweight basis for motion segmentation, moving-object detection, and temporally aware perception. Future work will focus on greater robustness under rotational motion and tighter integration with event-based detection and tracking.

\begingroup
\small
\bibliographystyle{splncs04}
\bibliography{references}
\endgroup

\end{document}